\pdfoutput=1
\PassOptionsToPackage{table}{xcolor}
\documentclass[11pt]{article}

\usepackage[preprint]{acl}

\usepackage{times}
\usepackage{latexsym}

\usepackage[T1]{fontenc}

\usepackage[utf8]{inputenc}

\usepackage{microtype}

\usepackage{inconsolata}

\usepackage{graphicx}
\usepackage{geometry}
\usepackage{xcolor}
\usepackage[table]{xcolor}
\usepackage[english]{babel}
\usepackage{tabularx}
\usepackage{multirow}
\usepackage{booktabs}
\usepackage{tcolorbox}
\usepackage{microtype}
\usepackage{url}
\usepackage{hyperref}
\usepackage{pifont}
\usepackage{amsmath}

\makeatletter
\g@addto@macro{\UrlBreaks}{\UrlOrds}
\makeatother

\newtcolorbox{prompt}[2][]{
fontupper = \linespread{.85}\fontsize{10pt}{12pt}\selectfont,
  colframe = gray,
  colback  = white,
  coltitle = #2!20!black,  
  segmentation style={solid},
  boxrule=1pt,
  left=2pt,
  right=2pt,
  top=2pt,
  bottom = 2pt,
  middle = 2pt,
  arc=0pt,
  #1,
}

\title{Beyond Translation: LLM-Based Data Generation for Multilingual Fact-Checking}

\author{Yi-Ling Chung, Aurora Cobo, Pablo Serna \\
        Genaios Safe AI
}

\begin{document}
\maketitle
\begin{abstract}
Robust automatic fact-checking systems have the potential to combat online misinformation at scale. However, most existing research primarily focuses on English. 
In this paper, we introduce \textbf{MultiSynFact}, the first large-scale multilingual fact-checking dataset containing 2.2M claim-source pairs designed to support Spanish, German, English, and other low-resource languages. Our dataset generation pipeline leverages Large Language Models (LLMs), integrating external knowledge from Wikipedia and incorporating rigorous claim validation steps to ensure data quality. We evaluate the effectiveness of MultiSynFact across multiple models and experimental settings. Additionally, we open-source a user-friendly framework to facilitate further research in multilingual fact-checking and dataset generation. 

\end{abstract}

\section{Introduction}
Online misinformation brings a major societal challenge.
Research has advanced in developing fact-checking resources \cite[e.g.][]{thorne-etal-2018-fever, NielsenMcConville2022} and automated solutions \cite{schuster-etal-2021-get, tianfine}, but most efforts focus on the English language. Multilingual fact-checking often relies on translation tools to generate datasets \cite[e.g.][]{shafayat2024multifact}, which, while effective, overlook geographic, cultural, and linguistic nuances \cite{guo-etal-2022-survey}. Moreover, human annotation offers high accuracy but remains costly and resource-intensive \cite{le2024viwikifc}.

In this context, LLMs have been widely explored for creating and enhancing training datasets \cite{long-etal-2024-llms, goyal2024systematic}. Prior work shows that models trained on synthetic data achieve performance comparable to those trained on real data \cite{li2023synthetic, tianfine}. While this approach holds great potential \cite{li2023synthetic, xu2024wizardlm}, two key areas remain for exploration: (1) its effectiveness in knowledge-intensive tasks like multilingual fact-checking and (2) the overall quality of generated data.  

\begin{table}[ht!]
\small
\begin{tabular}{p{0.95\linewidth}}
\toprule
\textbf{Source:} In J. R. R. Tolkien's legendarium, the Elves or Quendi are a sundered (divided) people. \\
\midrule
\textbf{Supports:} The Quendi, as described in J. R. R. Tolkien's 
legendarium, are a divided people.  \\
\midrule
\textbf{Refutes}: The Elves in Tolkien's legendarium are not sundered but united by common ancestry. \\
\midrule
\textbf{Not-info:} The Elves' sundering in Tolkien's legendarium may 
have been more profound than that of other fictional divided peoples. \\
\midrule
\textbf{Topic:} Sundering of the Elves.\\
\bottomrule
\end{tabular}
\caption{Generation example of claims for each class.}
  \label{table:example_claims}
\end{table}

\begin{figure*}[h!]
    \centering
    \includegraphics[width=0.80\textwidth]{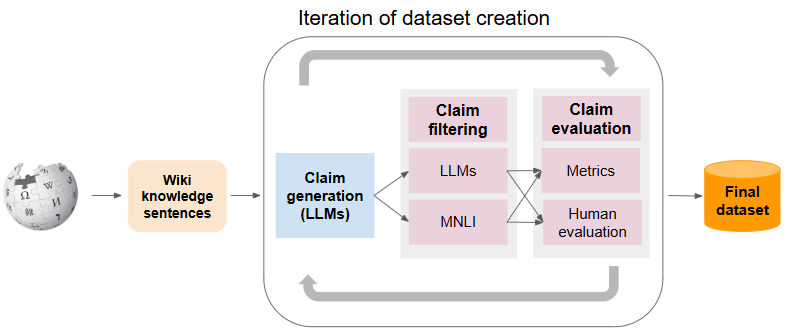}
    \caption{Automated pipeline of multilingual claims generation.}
    \label{fig:pipeline}
\end{figure*}

In this paper, we propose a scalable and efficient pipeline for automatically generating high-quality multilingual fact-checking datasets using Wikipedia as a source of knowledge (see Figure~\ref{fig:pipeline}), inspired by the FEVER dataset \cite{thorne-etal-2018-fever}. To our knowledge, this is the first work using LLMs for synthesising multilingual fact-checking data with external knowledge.

Our pipeline follows three steps: (1) extracting knowledge sentences from Wikipedia, (2) generating claims categorised as \textit{Supports}, \textit{Refutes}, or \textit{Not-info} (not enough information to decide) using LLMs, (3) applying rigorous validation to ensure linguistic and semantic alignment with the sources. This results in large-scale, multilingual and synthetic fact-checking datasets, and more specifically, we create and present MultiSynFact, containing 2.2M source-claim pairs across German, Spanish, and English languages (see English examples in Table \ref{table:example_claims} and Spanish and German examples in Appendix \ref{app:generation_examples}). This pipeline is scalable and adaptable to other languages, including low-resource ones, significantly reducing effort, costs and time. 

To assess the impact of our dataset, we conduct extensive experiments comparing fact-checking models trained on our synthetic data, with those trained on existing datasets in monolingual, multilingual and cross-lingual settings. Results consistently show that incorporating MultiSynFact during training enhances model generalisation, yielding higher macro F1 scores in nearly all cases, but especially in Spanish and German. 
The data and full implementation of our method will be available as an open-source toolkit at \url{https://github.com/Genaios/MultiSynFact}.

\section{Related Work}\label{sec:related_work}
\paragraph{Synthetic Data Generation.}
Synthetic data has been widely explored to alleviate data scarcity and quality challenges, enabling the development of robust models while significantly reducing costs and training time \cite{long-etal-2024-llms, goyal2024systematic, patel2024datadreamer}. In scientific domains, \citet{wright-etal-2022-generating} leveraged LLMs to generate supporting claims and create refuting claims through targeted replacements.  
The application of synthetic data spans various domains \cite{li2023synthetic, xu2024wizardlm, xu2024magpie}, with its impact on model classification performance varying depending on the task \cite{chan2024balancing}. Whilst prior studies suggest that verifying synthetic data does not always lead to direct performance gains \cite{li2023synthetic, yu2024metamath, chan2024balancing}, our work demonstrates the effectiveness of LLM-generated data in enhancing multilingual fact-checking. Specifically, we show that incorporating our synthetic data improves model generalisation across diverse linguistic settings.

\paragraph{Fact-Checking Datasets.}
While misinformation is a global issue affecting all languages, research on multilingual fact-checking datasets remains limited. Most existing datasets focus primarily on the English language \cite{thorne-etal-2018-fever, schuster-etal-2021-get}, with only a few covering other languages such as Spanish and German \cite{gupta2021x}, Danish \cite{norregaard-derczynski-2021-danfever}, or Arabic \cite{khouja-2020-stance}. 
Moreover, non-English language datasets are often small and domain-specific, primarily addressing topics such as COVID-19 \cite{li2020mmcovid, shahifakecovid}, social networks \cite{NielsenMcConville2022} and news \cite{gupta2021x}. 
To address this limitation, we introduce a scalable pipeline capable of generating large-scale multilingual fact-checking datasets for any language. Whilst our analysis focuses on Spanish, German, and English languages, our approach is inherently adaptable to a wide range of languages, ensuring its applicability in multilingual fact-checking. Table \ref{table:datasets_comparison} provides a comparative analysis of our dataset against existing multilingual and synthetic ones.

\begin{table*}[ht!]
\small
\centering
\resizebox{\textwidth}{!}{
\begin{tabular}{lrlclc}
\toprule
Datasets                                            &   Size & Source        & Annotation & Lang         & Synthetic\\
\midrule
ANS \cite{khouja-2020-stance}                       &  4,547 & news          & 2 Classes  & ar          & \ding{55}   \\
FakeCovid \cite{shahifakecovid}                    &  5,182 & fact-checking & 2 Classes  & Many        & \ding{55}   \\
MM-COVID \cite{li2020mmcovid}                       & 11,173 & Twitter/X     & 2 Classes  & Many        & \ding{55}   \\
X-Fact \cite{gupta2021x}                            & 31,189 & fact-checking & 7 Classes  & Many        & \ding{55}   \\
DanFEVER \cite{norregaard-derczynski-2021-danfever} &  6,407 & Wikipedia     & 3 Classes  & da          & \ding{55}   \\
ClaimGen \cite{wright-etal-2022-generating}        & 10,153 & science       & 2 Classes  & en          & \ding{51}   \\
MuMiN \cite{NielsenMcConville2022}                  & 12,914 & Twitter/X     & 2 Classes  & Many        &  \ding{55}  \\
MMFC \cite{bussotti-etal-2024-unknown}              & 1,500  & Wikipedia     & 2 Classes  & en          & \ding{51}   \\
\midrule
MultiSynFact (Ours                                                )&  2.2M  & Wikipedia     & 3 Classes  & en, de, es  & \ding{51}   \\
\bottomrule
\end{tabular}}
\caption{A comparison of multilingual or synthetic datasets for fact-checking.}
  \label{table:datasets_comparison}
\end{table*}

\paragraph{Fact-Checking Techniques.}
Methods for verifying claims against sources typically fall into two categories. The first approach involves developing specialised tools, primarily by fine-tuning pre-trained language models on labeled datasets \cite{thorne-etal-2018-fever, schuster-etal-2021-get, tianfine}. The second approach directly querying LLMs for factual evaluation without fine-tuning \cite{hu2024towards, shafayat2024multifact, 10.3389/frai.2024.1341697, singhal-etal-2024-multilingual}. 
While LLMs have shown promising results, their predictions often lack consistency across languages and claim veracity \cite{10.3389/frai.2024.1341697}. 
To enhance the factuality of LLMs, various techniques have been explored, including learning from automatically generated data, such as reference-free truthfulness estimation based on model confidence \cite{tianfine}, and instruct-tuning with external knowledge \cite{10317251}.

\section{Multilingual Dataset Generation}\label{sec:generation_pipeline}
Our pipeline for generating multilingual datasets, illustrated in Figure~\ref{fig:pipeline}, comprises four key components designed to prepare factual knowledge sources and generate and validate claims using LLMs. First, a \textit{knowledge sentences creation} component extracts and samples sentences from Wikipedia. Next, a \textit{claim generation} component produces claims based on the parsed sentences. 
Finally, \textit{claim filtering and evaluation} components validate the generated claims against a set of predefined metrics. 
The claim generation and validation processes can be refined iteratively to enable qualitative improvements over multiple iterations. Our final synthetic datasets were created after five iterations of generation and validation processes.

\subsection{Knowledge Sentences Creation}\label{subsec:senteces_creation}
The first step involves preparing factual knowledge sentences as sources. For this, we used the April 2024 Wikipedia dump (20240401) for Spanish and German languages and the August 2024 dump (20240820) for the English language, parsing the data using \textit{wikitextparser}\footnote{\url{https://pypi.org/project/wikitextparser/}}.
For each Wikipedia entry, we created two types of knowledge sentences to enhance data diversity by varying the sources. 
The first type of knowledge sentences comprises five sentences randomly sampled from a Wikipedia page. The second type includes three sentences--specifically, the first sentence, a randomly selected sentence, and the last sentence from the summary section. This approach yields a total of eight knowledge sentences per entry. 
Note that the automatically retrieved sources may sometimes be suboptimal, for instance, containing incomplete/ill-formatted sentences or lacking sufficient context. However, we consider this limitation to better reflect real-world scenarios, where claims are frequently incomplete or ambiguous \cite{10.1162/tacl_a_00629}.

\subsection{Claim Generation} \label{subsec:claim_gen}
After preparing the knowledge source sentences, we use \textit{Mistral-7B-Instruct-v0.3}\footnote{We refer Mistral-7B to Mistral-7B-Instruct-v0.3 throughout the paper.} \cite{jiang2023mistral} to generate claims for each of the three classes (with labels: \textit{supports}, \textit{refutes} and \textit{not-info}). This process is performed for each source sentence. Mistral-7B was chosen for its multilingual capabilities at the time of this research. The prompt used for generating claims with class supports is presented in Table \ref{table:support_prompt}. For the prompts for class refutes and not-info, refer to Appendix \ref{app:generation_prompts}. 

We randomly sample 30,000 Wikipedia entries  (i.e. 240,000 sentences) as sources for claim generation. To enhance the sensitivity of models to contrastive examples \cite{schuster-etal-2021-get}, we further instruct Mistral-7B to generate claims containing comparative or superlative adjectives (e.g., \textit{``X is larger than Y''}). Models trained on such variation are expected to better learn source-claim inference, enabling them to remain more faithful to facts or events that may evolve over time \cite{jacovi-goldberg-2020-towards, schuster-etal-2021-get}.

\begin{table*}[h!]
\centering
\begin{tabularx}{\textwidth}{|X|}
\hline
  Act as an expert generating claims in <language>. I will give you one evidence in <language> about the topic: "<topic>". Generate a single short and objective claim in <language>, with factual information about the sentence I will provide, using the information available in it. The claim should be supported by the evidence. Also, the claim should use comparative form (e.g., larger, smaller, more, less, faster, higher) to make comparisons between objects, people, ideas, dates or numbers in the evidence. The evidence is: "<sources>". Do not add any explanation, refrain from adding extra information nor your opinion on whether the claim is true or not, as long as it is supported by the evidence. The claim should have less than 30 words. Do not make any reference to the sentence in your answer, make it self-contained. Write your answer in {language} only. Evaluate at the end how good the generated claim is. Your response must be in the same format as the JSON in the examples below. \\

Your response must be in the same format as the JSON in the examples below. \\
\{\{    \\
\hspace{10mm} "CLAIM": "Write a single short and objective claim in <language>, with factual information about the sentence",  \\
\hspace{10mm} "SELF-CONTAINED": "Assess how self-contained the generated claim is on a scale of 1 to 5.",  \\
\hspace{10mm} "CATEGORY": "Categorise whether the given sentence not supported (C0), supported (C1) the generated claim (independently of actual veracity of the claim), or not verifiable (C2)",  \\
\hspace{10mm} "SUPPORTED BY ORIGINAL SENTENCE": "Assess how supported the claim is by the original sentence on a scale of 1 to 5.",  \\
\hspace{10mm} "FACTUAL": "Assess how factual the claim is based on the original sentence [real/non-fiction/non-fantastic]",  \\
\hspace{10mm} "OBJECTIVE": "Assess how objective the claim is on a scale of 1 to 5",  \\
\hspace{10mm} "OVERALL QUALITY": "Assess the overall quality of the claim on a scale of 1 to 5" \\
\}\}. \\
\hline
\end{tabularx}
\caption{The prompt for generating \textit{supporting} claims.}
  \label{table:support_prompt}
\end{table*}

\subsection{Claim Filtering} \label{subsec:claim_fil}
The next step involves developing a robust filtering mechanism to automatically evaluate the quality of generated claims.  
Given the limited research on multilingual fact-checking, our approach focuses on creating multiple filters to approximate the selection of high-quality claims while minimising the need for human intervention. 
To achieve this, we consider two types of reference-free criteria: (1) filters based on LLMs and (2) filters using Multilingual Natural Language Inference (MNLI). These criteria leverage the capabilities of advanced LLMs for handling knowledge-intensive tasks. 
We justify the use of the MNLI filtering in Appendix \ref{app:ablation}. Moreover, since our approach is based on multilingual models, it is applicable to all languages supported by these models without requiring additional fine-tuning. 

\paragraph{LLM.} We utilise the same LLM for claim generation to evaluate the quality of the generated claims across four aspects on a scale of 1 to 5: \textit{self-contained} (how well the claim stands independently without requiring additional context), \textit{support} (how well the claim is supported by the source sentences), \textit{objective} (how objective the claim is), and \textit{quality} (the overall quality of the claim). Additionally, the LLM is instructed to categorise the claims into three \textit{categories}: C0 (claims contradicted by the source sentences), C1 (claims supported by the source sentences), or C2 (claims that are unverifiable based on the source sentences). All classes are based on the sources, independently of their actual veracity. We also assess the \textit{factuality} of the claims, determining how factual, realistic, and non-fantastic they are based on the source sentences (real/non-fiction/non-fantastic). For the filtering process, we keep only the claims that are labeled with the same category as the target class (C0 for `refutes', C1 for `supports' and C2 for `not-info') and with scores above 3 in both the \textbf{quality} and \textbf{self-contained} aspects.

\paragraph{MNLI.} For this step, we use \textit{mDeBERTa-v3-base-xnli-multilingual-nli-2mil7} model\footnote{\url{https://huggingface.co/MoritzLaurer/mDeBERTa-v3-base-xnli-multilingual-nli-2mil7}}, a \textit{mDeBERTa-v3-base} model fine-tuned on the XNLI \cite{conneau2018xnli} and the multilingual-NLI-26lang-2mil7 \cite{laurer_less_2022} datasets. This model was selected because NLI and fact-checking are inherently similar tasks, both aiming to measure semantic congruence between two texts \cite{guo-etal-2022-survey}. We frame the source sentences as premises and the generated claims as hypotheses. During prediction, the model classifies the relationship between the premise and hypothesis into one of the three categories: \textbf{entailment} (interpreted as supporting the claim), \textbf{contradiction} (interpreted as refuting the claim) and \textbf{neutral} (interpreted as not-info). To ensure alignment with the target class, we discard any claims where the model's predictions do not match the intended class. This step ensures the filtered dataset maintains consistency and relevance to the task at hand. 

\subsection{Claim Evaluation} \label{subsec:claim_ev}
\paragraph{Automatic Evaluation.}
We assess the quality of generated claims based on lexical similarity between source sentences and claims via BLEU \cite{papineni2002bleu}, ROUGE \cite{lin-2004-rouge} and METEOR \cite{banerjee-lavie-2005-meteor} metrics. Specifically, we report BLEU-4 and ROUGEL.

\paragraph{Human Evaluation.}
To complement automatic evaluation, we also conduct human assessments to measure the quality of the generated claims and refine the prompts used for querying the LLM. For a given prompt, we randomly sample 10 claims for each class. Two authors of this paper evaluated these claims on a 1-5 scale based on: \textbf{overall quality} (the coherence and informativeness of the claims), \textbf{grammaticality} (the grammatical correctness of the claims) and \textbf{semantic relation} (the logical and factual relationship between the source sentences and the claims). 
They also verified whether the predicted labels were accurate. This iterative process of claim generation and validation continues until the generated claims achieve a score above 4 in all aspects.

\subsection{Dataset Analysis} \label{subsec:final_data}

Following the steps in Sections \ref{subsec:senteces_creation}-\ref{subsec:claim_ev}, we generate two datasets to support research in multilingual fact-checking and efficient data generation:

\begin{itemize}
    \item \textbf{Dataset without MNLI filtering} (\textit{no\_mnli\_filtering}): 3.8M instances filtered only with LLM methods.
    \item \textbf{Dataset with MNLI filtering} (\textit{mnli\_filtering}): 2.2M instances filtered using both LLM and MNLI methods.

\end{itemize}

As summarised in Table \ref{table:data}, the class distribution is skewed towards \textit{not-info} class, highlighting the inherent difficulty in generating refuting claims particularly. This imbalance aligns with the challenges identified in prior research  \cite{bilu-etal-2015-automatic} and underscores the need for strategies to enhance the generation of such claims.

Furthermore, we observe that the disagreement between MNLI model and LLM predictions is more pronounced for the \textit{refutes} class, while it remains relatively lower for the \textit{supports} and \textit{not-info} classes. We report the statistics and evaluation results of our dataset created with MNLI filtering using automatic metrics in Table \ref{table:statistics_results}. Overall we consider the generated claims in our dataset objective and self-contained, with scores above 4 per class and language. Moreover, the generated claims are lexically similar to the sources provided, in particular for the class \textit{supports}, based on BLEU-4, ROUGEL and METEOR.
The breakdown of statistics from automatic evaluation for our dataset without MNLI filtering is summarised in Appendix \ref{app:data_statistics}.

\begin{table}[ht!]
\small
\centering
\resizebox{\columnwidth}{!}{
\begin{tabular}{lrr}
\toprule
                        & MNLI filtering & no MNLI filtering \\
\midrule
EN                      &  895,223        &  1,521,694      \\
  \hspace{3mm} supports  &  264,460        &    469,582      \\
  \hspace{3mm} refutes  &  179,446        &    479,500      \\
  \hspace{3mm} not-info  &  451,317        &    572,612      \\
\midrule
ES                      &  700,268        &  1,152,701      \\
  \hspace{3mm} supports  &  253,285        &    397,184      \\
  \hspace{3mm} refutes  &  115,413        &    335,463      \\
  \hspace{3mm} not-info  &  331,570        &    420,054      \\
\midrule
DE                      &  680,034        &  1,168,905      \\
  \hspace{3mm} supports  &  214,651        &    357,069      \\
  \hspace{3mm} refutes  &  126,629        &    371,917      \\
  \hspace{3mm} not-info  &  338,754        &    439,919      \\
\bottomrule
\end{tabular}}
\caption{The size of data created with and without MNLI filtering.}
  \label{table:data}
\end{table}

\begin{table*}[!ht]
    \small
    \centering
    \resizebox{\textwidth}{!}{
    \begin{tabular}{lrrrrrrrrr}
    \toprule
                         & \multicolumn{2}{c}{Statistics}  & \multicolumn{4}{c}{LLM evaluation} & \multicolumn{3}{c}{Lexical similarity}\\
                           \cmidrule(lr){2-3}                \cmidrule(lr){4-7}      \cmidrule(lr){8-10}
          ~              &  words$_\mu$     & words$_\text{sd}$ & self-contained &	support	& objective &  quality &  BLEU-4 & ROUGEL & METEOR\\              
    \midrule
    EN                   &                  &              &                &         &           &         &         &        &     \\
    \hspace{3mm} supports &      16          &      6       &      5         &    5    &    5      &    4    &   0.22  &  0.50  &  0.49\\
    \hspace{3mm} refutes &      12          &      4       &      5         &    1    &    4      &    4    &   0.10  &  0.35  &  0.31\\
    \hspace{3mm} not-info &      15          &      4       &      5         &    2    &    4      &    4    &   0.06  &  0.24  &  0.24\\
    \midrule
    ES                   &                  &              &                &         &           &         &         &        &    \\
    \hspace{3mm} supports &      16          &      6       &      5         &    5    &    5      &    4    &   0.20  &  0.50  &  0.45\\
    \hspace{3mm} refutes &      12          &      4       &      5         &    1    &    4      &    4    &   0.08  &  0.34  &  0.28\\
    \hspace{3mm} not-info &      14          &      4       &      5         &    1    &    4      &    4    &   0.05  &  0.24  &  0.20\\
    \midrule
    DE                   &                  &              &                &         &           &         &         &        &    \\
    \hspace{3mm} supports &      13          &      5       &      5         &    5    &    5      &    4    &   0.21  &  0.50  &  0.45\\
    \hspace{3mm} refutes &      10          &      3       &      5         &    1    &    4      &    4    &   0.10  &  0.34  &  0.29\\
    \hspace{3mm} not-info &      12          &      4       &      5         &    1    &    4      &    4    &   0.05  &  0.20  &  0.18\\
    \bottomrule
    \end{tabular}}
    \caption{Statistics and evaluation of MultiSynFact, our dataset with MNLI filtering.  Columns $words_\mu$ and $words_\text{sd}$ show the average and standard deviation of the number of words per claim. The following columns show average values for the respective metric denoted as the column names. Lexical similarity metrics are computed between sources and generated claims.}
    \label{table:statistics_results}
\end{table*}

\section{Experimental Settings}
We evaluate whether incorporating our synthetic data (the one with the MNLI filtering from now on), MultiSynFact, into model training improves fact-checking performance compared to using evaluation dataset by itself. We consider three widely-used datasets in the fact-checking literature: X-Fact \cite{gupta2021x}, VitaminC \cite{schuster-etal-2021-get} and NLI-Fever \cite{nie2019combining}. X-Fact is a multilingual fact-checking dataset sourced from verified fact-checking websites, available for 25 languages. VitaminC is a widely used benchmark dataset for English language fact-checking tasks. NLI-Fever is a reformulated version of FEVER dataset designed in an NLI style. As VitaminC and NLI-Fever are available only in the English language, we translate them to Spanish and German using LibreTranslate\footnote{https://github.com/LibreTranslate/LibreTranslate}. 

All details in the pre-processing of the datasets are described in Appendix \ref{app:datasets_pre}.

For model training, we fine-tune the \textit{mDeBERTa-v3-base} model \cite{he2021debertav3} with a classifier head producing outputs for three classes using 1 NVIDIA H100 GPU.\footnote{We refer mDeBERTa-base to mDeBERTa-v3-base throughout the paper.} The training details are outlined in Appendix \ref{app:training_details}.

To analyse the impact of our synthetic data, we conduct experiments as follows:
\begin{enumerate}
    \item \textbf{Monolingual}: Training and testing in the same language.
    \item \textbf{Multilingual}: Training and testing across different languages seen together.
    \item \textbf{Cross-Lingual}: Training in Spanish, German and English languages together while testing on unseen languages.
\end{enumerate}

\paragraph{Baselines.} To establish benchmarks, we conduct the same training processes without incorporating our synthetic data. This allows us to measure the added value of the synthetic data.
Additionally, we evaluate GPT-4o \cite{GPT-4o}, GPT-4o-mini \cite{GPT-4o-mini}, Mistral-7B, Llama 3.3-70B \cite{touvron2023llama} and Phi4 \cite{abdin2024phi} on the test splits of X-fact, Vitamin-C and NLI-Fever datasets in zero-shot settings. These evaluations provide a comparative analysis of state-of-the-art LLMs for fact-checking tasks.

\section{Experimental Results}

\paragraph{Monolingual Results.}
In this section, we analyse the impact of incorporating our synthetic data on classification performance (in terms of macro F1 scores) compared to not using it. 
We present results for each dataset and language, distinguishing between originally sourced data and translations through our pipeline.
As shown in Table~\ref{tab:monolingual}, our synthetic data consistently boosts training, leading to improved F1 scores in all cases except VitaminC in English. We hypothesise that this may stem from VitaminC being used during the pretraining of \textit{mDeBERTa-base}, resulting in an already optimised performance. Given that its F1 score is the highest among all scenarios, it may also be more challenging to gain further improvements.

\begin{table}[h]
\small
\centering
\begin{tabular}{cl
>{\columncolor[HTML]{DAE9F8}}l 
>{\columncolor[HTML]{F2CEEF}}l }
\toprule
\cellcolor[HTML]{D0D0D0}\textbf{Data} & \multicolumn{1}{c}{\cellcolor[HTML]{D0D0D0}\textbf{lang}} & \cellcolor[HTML]{D0D0D0}\textbf{Without SD} & \cellcolor[HTML]{D0D0D0}\textbf{With SD} \\ 
\toprule
                                      & en                                                            & 0.750               & \textbf{0.757}   \\ \cline{2-4} 
                                      & es*                                                            & 0.730               & \textbf{0.734}   \\ \cline{2-4} 
\multirow{-3}{*}{NLI-Fever}          & de*                                                            & 0.725               & \textbf{0.734}   \\ 
\midrule
                                      & en                                                            & \textbf{0.851}      & 0.846            \\ \cline{2-4} 
                                      & es*                                                            & 0.819               & \textbf{0.825}   \\ \cline{2-4} 
\multirow{-3}{*}{VitaminC}            & de*                                                            & 0.814               & \textbf{0.819}   \\
\midrule
                                      & es                                                            & 0.262               & \textbf{0.336}   \\ \cline{2-4} 
\multirow{-2}{*}{X-Fact}             & de                                                            & 0.230               & \textbf{0.318}   \\ 
\bottomrule
\end{tabular}
\caption{Monolingual results of macro F1 scores. The train and test datasets are defined in the column `Data'. \textit{With SD} and \textit{without SD} indicate if the synthetic data was used for training or not, respectively. * denotes translated data.}
\label{tab:monolingual}
\end{table}

\paragraph{Multilingual Results.}
Table~\ref{tab:multilingual} presents the results for each dataset when training with the data in all available languages (25 languages in X-Fact and three in VitaminC and NLI-Fever), reporting the averaged F1 scores across the available target languages (Spanish and German for X-Fact and also English for VitaminC and NLI-Fever). 
Across all datasets, incorporating our synthetic data consistently improves performance, demonstrating the effectiveness of our pipeline in multilingual settings. Notably, these improvements extend beyond the English language, suggesting the robustness of our approach across diverse linguistic contexts.

\begin{table}[h]
\centering
\small
\begin{tabular}{l
>{\columncolor[HTML]{DAE9F8}}l 
>{\columncolor[HTML]{F2CEEF}}l}
\toprule
\cellcolor[HTML]{D0D0D0}\textbf{Data} & \cellcolor[HTML]{D0D0D0}\textbf{Without SD} & \cellcolor[HTML]{D0D0D0}\textbf{With SD}  \\
\midrule
NLI-Fever &   0.730 (0.012)   &  \textbf{0.742} (0.011)  \\
VitaminC &   0.836 (0.016)  &  \textbf{0.840} (0.015) \\
X-Fact &   0.308 (0.067)  &   \textbf{0.511} (0.219)    \\
\bottomrule
\end{tabular}
\caption{Multilingual results of macro F1 scores per dataset. Standard deviation is shown in parentheses for averaged results in testing languages.} 
\label{tab:multilingual}
\end{table}

\paragraph{Cross-Lingual Results.}

\begin{table*}[tb]
\small
\centering
\begin{tabular}{l|ccc|ccccccc}
\toprule
\textbf{Training Dataset} & \textbf{en} & \textbf{es} & \textbf{de} & \textbf{fr} & \textbf{it} & \textbf{bg} & \textbf{fi} & \textbf{zh} & \textbf{ja} & \textbf{hi} \\
\midrule
VitaminC + SD (es, en, de) & 0.846 & \textbf{0.825} & \textbf{0.819} &  \textbf{0.810}  &  \textbf{0.824} &  \textbf{0.787} &  \textbf{0.733} &  0.626  &  0.689  &  0.732  \\
VitaminC (es, en, de) & \textbf{0.851} & 0.819 & 0.814 &  0.805  &  0.818 &  0.786 &  0.728 &  \textbf{0.630}  &  \textbf{0.699}  &  \textbf{0.744}  \\
\bottomrule
\end{tabular}
\caption{Cross-lingual transfer performance (macro F1 scores) for unseen languages during training. This table shows the performance of mDeBERTa-base (fine-tuned on the corresponding training dataset shown in the first column) tested on translated versions of VitaminC in the languages specified in each column. Both VitaminC and the Synthetic dataset include English, Spanish and German (the latter two languages are translations in the case of VitaminC).}
\label{tab:cross_lingual}
\end{table*}

Table~\ref{tab:cross_lingual} presents the model's performance, tested on unseen languages during fine-tuning. We trained the model on the VitaminC dataset, both with and without our synthetic data. It was then tested on the translated VitaminC dataset in other languages, including out-of-scope and low-resource ones.
Our results confirm that incorporating synthetic data improves classification performance, particularly for languages similar to those seen during training. When introducing our synthetic data, we observe performance gains in Spanish, German, and Latin-based languages, with macro F1 scores above 0.8 for French and Italian. 
For languages linguistically distant from the training set, the performance drops more noticeably. However, the model demonstrates strong generalisability, achieving reasonable performance of F1 scores ranging between 0.6 and 0.8.

\paragraph{Cross-Model Comparison Results.}
As a final set of experiments, we report the results of comparing our model with five powerful LLMs (GPT-4o, GPT-4o-mini, Mistral-7B, Llama3.3 and Phi4). 
In this evaluation, we down-sampled dataset sizes to 10,000 when necessary and included all 25 languages in the original X-Fact dataset. The prompt used for the evaluation is shown in Appendix \ref{app:Cross-model-eval}. 
Results in Table~\ref{tab:cross_model} show that our model outperforms LLMs across all scenarios. Moreover, GPT models perform poorly on the X-Fact dataset in comparison to NLI-Fever and VitaminC datasets. This demonstrates that fact-checking remains challenging for LLMs and that developing specialised models remains a better strategy for fact verification. 

We further expand the cross-model evaluation of fact-checking by looking at the model performance in the different languages included in X-Fact. As shown in Table~\ref{tab:x_fact_languages}, adding the synthetic dataset (including Spanish, English and German) to the fine-tuned \textit{mdeberta-base} improves the performance of the model in Spanish and closely related languages such as Portuguese or Italian. In this particular case, it did not improve in German, although it did not hurt the performance either. We hypothesise that this is because the German data in X-Fact covers different topics from our MultiSynFact. Future work can explore methods for improvements in specific target languages. In general, training with MultiSynFact enhances the overall performance with respect to not using it and to the LLMs we tested. We report further details of model comparison across three transformer-based models in Appendix \ref{app:transformer_comparison}.

\begin{table*}[h!]
\small
\centering
\begin{tabular}{cl
>{\columncolor[HTML]{F2CEEF}}l 
>{\columncolor[HTML]{DAF2D0}}l 
>{\columncolor[HTML]{DAF2D0}}l 
>{\columncolor[HTML]{DAF2D0}}l 
>{\columncolor[HTML]{DAF2D0}}l 
>{\columncolor[HTML]{DAF2D0}}l }
\toprule
\cellcolor[HTML]{D0D0D0}\textbf{Data} & \multicolumn{1}{c}{\cellcolor[HTML]{D0D0D0}\textbf{Language}} & \cellcolor[HTML]{D0D0D0}\textbf{With SD} & \cellcolor[HTML]{D0D0D0}\textbf{GPT-4o} & \cellcolor[HTML]{D0D0D0}\textbf{GPT-4o-mini} & \cellcolor[HTML]{D0D0D0}\textbf{Mistral-7B} & \cellcolor[HTML]{D0D0D0}\textbf{Llama3.3}& \cellcolor[HTML]{D0D0D0}\textbf{Phi4} \\
\midrule
                                      & en                                    & \textbf{0.765}   & 0.731           & 0.657                &           0.510          &         0.696  & 0.654     \\
                                      &es                                    & \textbf{0.740}   & 0.671           & 0.606                &          0.450           &      0.653   & 0.620      \\
\multirow{-3}{*}{NLI-Fever 10k}      & de                                    & \textbf{0.741}   & 0.663           & 0.598                &        0.463             &     0.650    & 0.603       \\
\midrule
                                      & en                                    & \textbf{0.841}   & 0.759           & 0.685                &              0.450        &     0.722   &0.677       \\
                                      & es                                                            & \textbf{0.817}   & 0.728           & 0.662                &   0.422                  &     0.692 & 0.622          \\

\multirow{-3}{*}{VitaminC 10k}        & de                                                            & \textbf{0.815}   & 0.714           & 0.640                 &            0.389         &           0.682  & 0.607   \\
\midrule
X-Fact                               & all                                                           & \textbf{0.532}   & 0.374           & 0.354                &        0.280             &      0.356   & 0.373      \\
\bottomrule
\end{tabular}
\caption{Cross-model comparison of fact-checking performance (macro F1 scores) across datasets and languages. We employ the same test split of all datasets but down-sampling to 10,000 inputs for NLI-Fever and VitaminC. For X-Fact, `all' refers to all 25 languages in the original dataset. The third (pink) column represents the performance of mDeBERTa-base fine-tuned with MultiSynFact and evaluated in the corresponding dataset (denoted in the first column). The following columns (green) represent the performance of zero-shot classification using the respective LLMs as denoted by the column name.}
\label{tab:cross_model}
\end{table*}

\begin{table*}[h!]
\small
\centering
\begin{tabular}{c>{\columncolor[HTML]{F2CEEF}}c
>{\columncolor[HTML]{F2CEEF}}c 
>{\columncolor[HTML]{DAF2D0}}c
>{\columncolor[HTML]{DAF2D0}}c
>{\columncolor[HTML]{DAF2D0}}c
>{\columncolor[HTML]{DAF2D0}}c
>{\columncolor[HTML]{DAF2D0}}c}
\toprule
\cellcolor[HTML]{D0D0D0}\textbf{Language} & \cellcolor[HTML]{D0D0D0}\textbf{X-Fact} & \cellcolor[HTML]{D0D0D0}\textbf{X-Fact + SD} & \cellcolor[HTML]{D0D0D0}\textbf{GPT-4o} & \cellcolor[HTML]{D0D0D0}\textbf{GPT-4o-mini} & \cellcolor[HTML]{D0D0D0}\textbf{Mistral-7B} & \cellcolor[HTML]{D0D0D0}\textbf{Llama3.3} & \cellcolor[HTML]{D0D0D0}\textbf{Phi4} \\
\midrule
ar & \textbf{0.396} & 0.379 & 0.301 & 0.314 & 0.233 & 0.318 & 0.312 \\
de & 0.356 & 0.356 & \textbf{0.399} & 0.364 & 0.313 & 0.330 & 0.342 \\
es & 0.261 & \textbf{0.666} & 0.519 & 0.461 & 0.364 & 0.467 & 0.439 \\
hi & \textbf{0.392} & 0.360 & 0.322 & 0.208 & 0.354 & 0.240 & 0.282 \\
id & 0.381 & \textbf{0.441} & 0.334 & 0.368 & 0.315 & 0.345 & 0.329 \\
it & 0.574 & \textbf{0.589} & 0.305 & 0.373 & 0.160 & 0.373 & 0.323 \\
ka & 0.301 & \textbf{0.368} & 0.359 & 0.280 & 0.317 & 0.301 & 0.378 \\
pl & 0.326 & 0.355 & 0.378 & 0.383 & 0.160 & 0.375 & \textbf{0.400} \\
pt & 0.458 & \textbf{0.514} & 0.349 & 0.317 & 0.244 & 0.321 & 0.342 \\
ro & \textbf{0.380} & 0.355 & 0.296 & 0.276 & 0.272 & 0.275 & 0.269 \\
sr & 0.393 & \textbf{0.418} & 0.219 & 0.173 & 0.248 & 0.160 & 0.264 \\
ta & \textbf{0.644} & 0.632 & 0.267 & 0.202 & 0.248 & 0.268 & 0.294 \\
tr & 0.329 & 0.397 & \textbf{0.443} & 0.410 & 0.179 & 0.395 & 0.409 \\
\midrule
AVG & 0.399 & \textbf{0.449} & 0.345 & 0.318 & 0.262 & 0.321 & 0.337 \\
\bottomrule
\end{tabular}
\caption{Macro F1 scores for cross-model comparison of fact-checking performance across all languages included in X-Fact. The second and third (pink) columns represent the performance of mDeBERTa-base, which has been fine-tuned on the datasets specified in their names (X-Fact and X-Fact + SD). The following columns (green) indicate the performance of zero-shot classification attempts using the respective LLMs as denoted by the column headers. The last row shows the average value.}\label{tab:x_fact_languages}
\end{table*}

\section{Discussion and Conclusion}
This paper tackles the problem of data scarcity for online misinformation in multilingual scenarios. We present an automatic pipeline to curate and validate multilingual claims grounded on Wikipedia. Leveraging LLMs and automated filtering techniques, we release 2.2M claim-source pairs for Spanish, German, and English languages to facilitate and further fact-checking research.
Through our experiments, we observe consistent improvements in all tested datasets after introducing our synthetic data. This confirms that our generation pipeline with filtering strategies generates high-quality claims, bringing a step towards developing robust fact-checking models. Moreover, the impact of the synthetic data is more evident when training data is limited - we observe more prominent F1 improvement in X-Fact (1,784 instances on average per language) than in NLI-Fever and VitaminC (at least 144k instances in the training set). 

Our experiments help understand the impact of synthetic data, providing a strong baseline for multilingual fact-checking, even for unseen languages. As our generation pipeline requires little human intervention, it can be easily deployed for generating diverse and novel claims with up-to-date evidence.
The promising results highlight future directions for developing fact-checking systems for low-resource languages.

While our pipeline is designed to support any language, it is bound by the language capability of the LLMs employed. For instance, for high-resource languages such as Italian and French, similar performance could be achieved by adopting the same approach and models used in the paper. For languages not supported by Mistral-7B, such as Arabic and Chinese, consider using Qwen \cite{qwen}. However, for low-resource languages, we expect that a more fine-grained analysis through multiple iterations is needed, including testing and identifying the best-performing LLMs for composing and validating claims. 

As future work, we plan to support more languages. We also plan to explore methods for further improving fact-checking systems, by generating diverse claims (both similar and dissimilar to current benchmarks) via instruction-tuning.

\section*{Limitations}
Our work addresses the lack of resources for developing multilingual fact-checking systems by generating synthetic datasets while reducing human effort. However, it faces several limitations. 

Firstly, our claims are generated based on sentences retrieved from Wikipedia, similar to previous work on dataset creation for fact-checking \cite{thorne-etal-2018-fever, norregaard-derczynski-2021-danfever}. Whilst Wikipedia is considered a public trustworthy source with reliable information and covers a broad range of topics and domains, it may contain unintended biases or false/undesired content (arising from the `wisdom of the crowd') or information about individual names available on Wikipedia. Such information may be reflected in the dataset. Adding diverse resources (e.g., news and fact-checking websites validated by journalists or human fact-checkers) could reduce potential biases and ensure the veracity of claims generated. Alternatively, applying toxicity classifiers could help identify potential offensive content to be removed. 

Secondly, as mentioned in Section \ref{subsec:senteces_creation}, our automatic source retrieval creates suboptimal factual sentences for claim generation. This may pose challenges in training robust fact-checking systems. 
Finally, we use a specific LLM (Mistral-7B) to generate claims due to its multilingual capability. The findings may not be generalised across models.  
We conducted an extensive evaluation comparing models trained on our dataset with existing real datasets to understand the robustness of our approach.

\section*{Ethical Considerations}
Tackling misinformation is an urgent task. As our paper aims to provide tools for efficiently creating high-quality fact-checking data for multiple languages using AI techniques, several societal and ethical consequences should be carefully managed. 

While our approach based on language models can scale up quickly across domains and languages, they can fail, such as making incorrect predictions, generating unfaithful texts, and potentially inducing unintended social biases (in both generation and classification tasks). These issues can be mitigated by, for instance, measuring the factual consistency between generated claims against retrieved sources \cite{min-etal-2023-factscore, zha-etal-2023-alignscore} and adding bias triggers into prompts for generation \cite{sheng-etal-2020-towards}.
Furthermore, as a first step towards creating multilingual synthetic datasets for fact-checking, we tested our pipeline for Spanish, German, and English, which are high-resource languages. It is crucial to investigate the effectiveness of our approach for low-resource scenarios for future work.
To facilitate research on multilingual fact-checking, our data and implementation are available under the license CC BY-NC 4.0.

\section*{Acknowledgments}
We are grateful to Stuart Winter-Tear for his valuable feedback on this paper. We thank Marc Franco-Salvador and José A. González-Barba for the discussions and encouragement in forming the paper.

\bibliography{ref}

\begin{thebibliography}{45}
\providecommand{\natexlab}[1]{#1}

\bibitem[{Abdin et~al.(2024)Abdin, Aneja, Behl, Bubeck, Eldan, Gunasekar, Harrison, Hewett, Javaheripi, Kauffmann et~al.}]{abdin2024phi}
Marah Abdin, Jyoti Aneja, Harkirat Behl, S{\'e}bastien Bubeck, Ronen Eldan, Suriya Gunasekar, Michael Harrison, Russell~J Hewett, Mojan Javaheripi, Piero Kauffmann, et~al. 2024.
\newblock Phi-4 technical report.
\newblock \emph{arXiv preprint arXiv:2412.08905}.

\bibitem[{Bai et~al.(2023)Bai, Bai, Chu, Cui, Dang, Deng, Fan, Ge, Han, Huang, Hui, Ji, Li, Lin, Lin, Liu, Liu, Lu, Lu, Ma, Men, Ren, Ren, Tan, Tan, Tu, Wang, Wang, Wang, Wu, Xu, Xu, Yang, Yang, Yang, Yang, Yao, Yu, Yuan, Yuan, Zhang, Zhang, Zhang, Zhang, Zhou, Zhou, Zhou, and Zhu}]{qwen}
Jinze Bai, Shuai Bai, Yunfei Chu, Zeyu Cui, Kai Dang, Xiaodong Deng, Yang Fan, Wenbin Ge, Yu~Han, Fei Huang, Binyuan Hui, Luo Ji, Mei Li, Junyang Lin, Runji Lin, Dayiheng Liu, Gao Liu, Chengqiang Lu, Keming Lu, Jianxin Ma, Rui Men, Xingzhang Ren, Xuancheng Ren, Chuanqi Tan, Sinan Tan, Jianhong Tu, Peng Wang, Shijie Wang, Wei Wang, Shengguang Wu, Benfeng Xu, Jin Xu, An~Yang, Hao Yang, Jian Yang, Shusheng Yang, Yang Yao, Bowen Yu, Hongyi Yuan, Zheng Yuan, Jianwei Zhang, Xingxuan Zhang, Yichang Zhang, Zhenru Zhang, Chang Zhou, Jingren Zhou, Xiaohuan Zhou, and Tianhang Zhu. 2023.
\newblock Qwen technical report.
\newblock \emph{arXiv preprint arXiv:2309.16609}.

\bibitem[{Banerjee and Lavie(2005)}]{banerjee-lavie-2005-meteor}
Satanjeev Banerjee and Alon Lavie. 2005.
\newblock \href {https://aclanthology.org/W05-0909} {{METEOR}: An automatic metric for {MT} evaluation with improved correlation with human judgments}.
\newblock In \emph{Proceedings of the {ACL} Workshop on Intrinsic and Extrinsic Evaluation Measures for Machine Translation and/or Summarization}, pages 65--72, Ann Arbor, Michigan. Association for Computational Linguistics.

\bibitem[{Bilu et~al.(2015)Bilu, Hershcovich, and Slonim}]{bilu-etal-2015-automatic}
Yonatan Bilu, Daniel Hershcovich, and Noam Slonim. 2015.
\newblock \href {https://doi.org/10.3115/v1/W15-0511} {Automatic claim negation: Why, how and when}.
\newblock In \emph{Proceedings of the 2nd Workshop on Argumentation Mining}, pages 84--93, Denver, CO. Association for Computational Linguistics.

\bibitem[{Bussotti et~al.(2024)Bussotti, Ragazzi, Frisoni, Moro, and Papotti}]{bussotti-etal-2024-unknown}
Jean-Flavien Bussotti, Luca Ragazzi, Giacomo Frisoni, Gianluca Moro, and Paolo Papotti. 2024.
\newblock \href {https://doi.org/10.18653/v1/2024.emnlp-main.675} {Unknown claims: Generation of fact-checking training examples from unstructured and structured data}.
\newblock In \emph{Proceedings of the 2024 Conference on Empirical Methods in Natural Language Processing}, pages 12105--12122, Miami, Florida, USA. Association for Computational Linguistics.

\bibitem[{Chan et~al.(2024)Chan, Pu, Shanker, Suresh, Jenks, Heyer, and Denton}]{chan2024balancing}
Yung-Chieh Chan, George Pu, Apaar Shanker, Parth Suresh, Penn Jenks, John Heyer, and Samuel~Marc Denton. 2024.
\newblock \href {https://openreview.net/forum?id=hRjFiTxv1v} {Balancing cost and effectiveness of synthetic data generation strategies for {LLM}s}.
\newblock In \emph{NeurIPS 2024 Workshop on Fine-Tuning in Modern Machine Learning: Principles and Scalability}.

\bibitem[{Cheung and Lam(2023)}]{10317251}
Tsun-Hin Cheung and Kin-Man Lam. 2023.
\newblock \href {https://doi.org/10.1109/APSIPAASC58517.2023.10317251} {Factllama: Optimizing instruction-following language models with external knowledge for automated fact-checking}.
\newblock In \emph{2023 Asia Pacific Signal and Information Processing Association Annual Summit and Conference (APSIPA ASC)}, pages 846--853.

\bibitem[{Conneau et~al.(2018)Conneau, Rinott, Lample, Williams, Bowman, Schwenk, and Stoyanov}]{conneau2018xnli}
Alexis Conneau, Ruty Rinott, Guillaume Lample, Adina Williams, Samuel~R. Bowman, Holger Schwenk, and Veselin Stoyanov. 2018.
\newblock Xnli: Evaluating cross-lingual sentence representations.
\newblock In \emph{Proceedings of the 2018 Conference on Empirical Methods in Natural Language Processing}. Association for Computational Linguistics.

\bibitem[{Glockner et~al.(2024)Glockner, Staliūnaitė, Thorne, Vallejo, Vlachos, and Gurevych}]{10.1162/tacl_a_00629}
Max Glockner, Ieva Staliūnaitė, James Thorne, Gisela Vallejo, Andreas Vlachos, and Iryna Gurevych. 2024.
\newblock \href {https://doi.org/10.1162/tacl_a_00629} {Ambifc: Fact-checking ambiguous claims with evidence}.
\newblock \emph{Transactions of the Association for Computational Linguistics}, 12:1--18.

\bibitem[{Goyal and Mahmoud(2024)}]{goyal2024systematic}
Mandeep Goyal and Qusay~H Mahmoud. 2024.
\newblock A systematic review of synthetic data generation techniques using generative ai.
\newblock \emph{Electronics}, 13(17):3509.

\bibitem[{Guo et~al.(2022)Guo, Schlichtkrull, and Vlachos}]{guo-etal-2022-survey}
Zhijiang Guo, Michael Schlichtkrull, and Andreas Vlachos. 2022.
\newblock \href {https://doi.org/10.1162/tacl_a_00454} {A survey on automated fact-checking}.
\newblock \emph{Transactions of the Association for Computational Linguistics}, 10:178--206.

\bibitem[{Gupta and Srikumar(2021)}]{gupta2021x}
Ashim Gupta and Vivek Srikumar. 2021.
\newblock X-fact: A new benchmark dataset for multilingual fact checking.
\newblock In \emph{Proceedings of the 59th Annual Meeting of the Association for Computational Linguistics and the 11th International Joint Conference on Natural Language Processing (Volume 2: Short Papers)}, pages 675--682.

\bibitem[{He et~al.(2021)He, Gao, and Chen}]{he2021debertav3}
Pengcheng He, Jianfeng Gao, and Weizhu Chen. 2021.
\newblock \href {https://arxiv.org/abs/2111.09543} {Debertav3: Improving deberta using electra-style pre-training with gradient-disentangled embedding sharing}.
\newblock \emph{Preprint}, arXiv:2111.09543.

\bibitem[{Hu et~al.(2024)Hu, Chen, Li, Guo, Wen, Yu, and Guo}]{hu2024towards}
Xuming Hu, Junzhe Chen, Xiaochuan Li, Yufei Guo, Lijie Wen, Philip~S. Yu, and Zhijiang Guo. 2024.
\newblock \href {https://openreview.net/forum?id=9OevMUdods} {Towards understanding factual knowledge of large language models}.
\newblock In \emph{The Twelfth International Conference on Learning Representations}.

\bibitem[{Jacovi and Goldberg(2020)}]{jacovi-goldberg-2020-towards}
Alon Jacovi and Yoav Goldberg. 2020.
\newblock \href {https://doi.org/10.18653/v1/2020.acl-main.386} {Towards faithfully interpretable {NLP} systems: How should we define and evaluate faithfulness?}
\newblock In \emph{Proceedings of the 58th Annual Meeting of the Association for Computational Linguistics}, pages 4198--4205, Online. Association for Computational Linguistics.

\bibitem[{Jiang et~al.(2023)Jiang, Sablayrolles, Mensch, Bamford, Chaplot, Casas, Bressand, Lengyel, Lample, Saulnier et~al.}]{jiang2023mistral}
Albert~Q Jiang, Alexandre Sablayrolles, Arthur Mensch, Chris Bamford, Devendra~Singh Chaplot, Diego de~las Casas, Florian Bressand, Gianna Lengyel, Guillaume Lample, Lucile Saulnier, et~al. 2023.
\newblock Mistral 7b.
\newblock \emph{arXiv preprint arXiv:2310.06825}.

\bibitem[{Khouja(2020)}]{khouja-2020-stance}
Jude Khouja. 2020.
\newblock \href {https://doi.org/10.18653/v1/2020.fever-1.2} {Stance prediction and claim verification: An {A}rabic perspective}.
\newblock In \emph{Proceedings of the Third Workshop on Fact Extraction and VERification (FEVER)}, pages 8--17, Online. Association for Computational Linguistics.

\bibitem[{Laurer et~al.(2022)Laurer, Atteveldt, Casas, and Welbers}]{laurer_less_2022}
Moritz Laurer, Wouter~van Atteveldt, Andreu~Salleras Casas, and Kasper Welbers. 2022.
\newblock \href {https://osf.io/74b8k} {Less {Annotating}, {More} {Classifying} – {Addressing} the {Data} {Scarcity} {Issue} of {Supervised} {Machine} {Learning} with {Deep} {Transfer} {Learning} and {BERT} - {NLI}}.
\newblock \emph{Preprint}.
\newblock Publisher: Open Science Framework.

\bibitem[{Le et~al.(2024)Le, To, Nguyen, and Van~Nguyen}]{le2024viwikifc}
Hung~Tuan Le, Long~Truong To, Manh~Trong Nguyen, and Kiet Van~Nguyen. 2024.
\newblock Viwikifc: Fact-checking for vietnamese wikipedia-based textual knowledge source.
\newblock \emph{arXiv preprint arXiv:2405.07615}.

\bibitem[{Li et~al.(2020)Li, Jiang, Shu, and Liu}]{li2020mmcovid}
Yichuan Li, Bohan Jiang, Kai Shu, and Huan Liu. 2020.
\newblock \href {https://arxiv.org/abs/2011.04088} {Mm-covid: A multilingual and multimodal data repository for combating covid-19 disinformation}.
\newblock \emph{Preprint}, arXiv:2011.04088.

\bibitem[{Li et~al.(2023)Li, Zhu, Lu, and Yin}]{li2023synthetic}
Zhuoyan Li, Hangxiao Zhu, Zhuoran Lu, and Ming Yin. 2023.
\newblock \href {https://openreview.net/forum?id=MmBjKmHIND} {Synthetic data generation with large language models for text classification: Potential and limitations}.
\newblock In \emph{The 2023 Conference on Empirical Methods in Natural Language Processing}.

\bibitem[{Lin(2004)}]{lin-2004-rouge}
Chin-Yew Lin. 2004.
\newblock \href {https://aclanthology.org/W04-1013} {{ROUGE}: A package for automatic evaluation of summaries}.
\newblock In \emph{Text Summarization Branches Out}, pages 74--81, Barcelona, Spain. Association for Computational Linguistics.

\bibitem[{Long et~al.(2024)Long, Wang, Xiao, Zhao, Ding, Chen, and Wang}]{long-etal-2024-llms}
Lin Long, Rui Wang, Ruixuan Xiao, Junbo Zhao, Xiao Ding, Gang Chen, and Haobo Wang. 2024.
\newblock \href {https://doi.org/10.18653/v1/2024.findings-acl.658} {On {LLM}s-driven synthetic data generation, curation, and evaluation: A survey}.
\newblock In \emph{Findings of the Association for Computational Linguistics: ACL 2024}, pages 11065--11082, Bangkok, Thailand. Association for Computational Linguistics.

\bibitem[{Min et~al.(2023)Min, Krishna, Lyu, Lewis, Yih, Koh, Iyyer, Zettlemoyer, and Hajishirzi}]{min-etal-2023-factscore}
Sewon Min, Kalpesh Krishna, Xinxi Lyu, Mike Lewis, Wen-tau Yih, Pang Koh, Mohit Iyyer, Luke Zettlemoyer, and Hannaneh Hajishirzi. 2023.
\newblock \href {https://doi.org/10.18653/v1/2023.emnlp-main.741} {{FA}ct{S}core: Fine-grained atomic evaluation of factual precision in long form text generation}.
\newblock In \emph{Proceedings of the 2023 Conference on Empirical Methods in Natural Language Processing}, pages 12076--12100, Singapore. Association for Computational Linguistics.

\bibitem[{Nie et~al.(2019)Nie, Chen, and Bansal}]{nie2019combining}
Yixin Nie, Haonan Chen, and Mohit Bansal. 2019.
\newblock Combining fact extraction and verification with neural semantic matching networks.
\newblock In \emph{Association for the Advancement of Artificial Intelligence ({AAAI})}.

\bibitem[{Nielsen and McConville(2022)}]{NielsenMcConville2022}
Dan~Saattrup Nielsen and Ryan McConville. 2022.
\newblock \href {https://arxiv.org/abs/2202.11684} {Mumin: A large-scale multilingual multimodal fact-checked misinformation social network dataset}.
\newblock In \emph{Proceedings of the 45th International ACM SIGIR Conference on Research and Development in Information Retrieval (SIGIR)}. ACM.

\bibitem[{N{\o}rregaard and Derczynski(2021)}]{norregaard-derczynski-2021-danfever}
Jeppe N{\o}rregaard and Leon Derczynski. 2021.
\newblock \href {https://aclanthology.org/2021.nodalida-main.47/} {{D}an{FEVER}: claim verification dataset for {D}anish}.
\newblock In \emph{Proceedings of the 23rd Nordic Conference on Computational Linguistics (NoDaLiDa)}, pages 422--428, Reykjavik, Iceland (Online). Link{\"o}ping University Electronic Press, Sweden.

\bibitem[{{Open AI}(2025{\natexlab{a}})}]{GPT-4o-mini}
{Open AI}. 2025{\natexlab{a}}.
\newblock \href {https://openai.com/index/gpt-4o-mini-advancing-cost-efficient-intelligence/} {Gpt-4o mini: advancing cost-efficient intelligence}.
\newblock Accessed: 2 April 2025.

\bibitem[{{Open AI}(2025{\natexlab{b}})}]{GPT-4o}
{Open AI}. 2025{\natexlab{b}}.
\newblock \href {https://openai.com/index/hello-gpt-4o/} {Hello gpt-4o.}
\newblock Accessed: 2 April 2025.

\bibitem[{Papineni et~al.(2002)Papineni, Roukos, Ward, and Zhu}]{papineni2002bleu}
Kishore Papineni, Salim Roukos, Todd Ward, and Wei-Jing Zhu. 2002.
\newblock Bleu: a method for automatic evaluation of machine translation.
\newblock In \emph{Proceedings of the 40th annual meeting on association for computational linguistics}, pages 311--318. Association for Computational Linguistics.

\bibitem[{Patel et~al.(2024)Patel, Raffel, and Callison-Burch}]{patel2024datadreamer}
Ajay Patel, Colin Raffel, and Chris Callison-Burch. 2024.
\newblock \href {https://doi.org/10.18653/v1/2024.acl-long.208} {{D}ata{D}reamer: A tool for synthetic data generation and reproducible {LLM} workflows}.
\newblock In \emph{Proceedings of the 62nd Annual Meeting of the Association for Computational Linguistics (Volume 1: Long Papers)}, pages 3781--3799, Bangkok, Thailand. Association for Computational Linguistics.

\bibitem[{Quelle and Bovet(2024)}]{10.3389/frai.2024.1341697}
Dorian Quelle and Alexandre Bovet. 2024.
\newblock \href {https://doi.org/10.3389/frai.2024.1341697} {The perils and promises of fact-checking with large language models}.
\newblock \emph{Frontiers in Artificial Intelligence}, 7.

\bibitem[{Schuster et~al.(2021)Schuster, Fisch, and Barzilay}]{schuster-etal-2021-get}
Tal Schuster, Adam Fisch, and Regina Barzilay. 2021.
\newblock \href {https://doi.org/10.18653/v1/2021.naacl-main.52} {Get your vitamin {C}! robust fact verification with contrastive evidence}.
\newblock In \emph{Proceedings of the 2021 Conference of the North American Chapter of the Association for Computational Linguistics: Human Language Technologies}, pages 624--643, Online. Association for Computational Linguistics.

\bibitem[{Shafayat et~al.(2024)Shafayat, Kim, Oh, and Oh}]{shafayat2024multifact}
Sheikh Shafayat, Eunsu Kim, Juhyun Oh, and Alice Oh. 2024.
\newblock \href {https://openreview.net/forum?id=lkrH6ovzsj} {Multi-{FA}ct: Assessing factuality of multilingual {LLM}s using {FA}ctscore}.
\newblock In \emph{First Conference on Language Modeling}.

\bibitem[{Shahi and Nandini(2020)}]{shahifakecovid}
Gautam~Kishore Shahi and Durgesh Nandini. 2020.
\newblock \href {http://workshop-proceedings.icwsm.org/pdf/2020_14.pdf} {Fake{C}ovid -- a multilingual cross-domain fact check news dataset for covid-19}.
\newblock In \emph{Workshop Proceedings of the 14th International {AAAI} {C}onference on {W}eb and {S}ocial {M}edia}.

\bibitem[{Sheng et~al.(2020)Sheng, Chang, Natarajan, and Peng}]{sheng-etal-2020-towards}
Emily Sheng, Kai-Wei Chang, Prem Natarajan, and Nanyun Peng. 2020.
\newblock \href {https://doi.org/10.18653/v1/2020.findings-emnlp.291} {Towards {C}ontrollable {B}iases in {L}anguage {G}eneration}.
\newblock In \emph{Findings of the Association for Computational Linguistics: EMNLP 2020}, pages 3239--3254, Online. Association for Computational Linguistics.

\bibitem[{Singhal et~al.(2024)Singhal, Law, Kassner, Gupta, Duan, Damle, and Li}]{singhal-etal-2024-multilingual}
Aryan Singhal, Thomas Law, Coby Kassner, Ayushman Gupta, Evan Duan, Aviral Damle, and Ryan~Luo Li. 2024.
\newblock \href {https://doi.org/10.18653/v1/2024.nlp4pi-1.2} {Multilingual fact-checking using {LLM}s}.
\newblock In \emph{Proceedings of the Third Workshop on NLP for Positive Impact}, pages 13--31, Miami, Florida, USA. Association for Computational Linguistics.

\bibitem[{Thorne et~al.(2018)Thorne, Vlachos, Christodoulopoulos, and Mittal}]{thorne-etal-2018-fever}
James Thorne, Andreas Vlachos, Christos Christodoulopoulos, and Arpit Mittal. 2018.
\newblock \href {https://doi.org/10.18653/v1/N18-1074} {{FEVER}: a large-scale dataset for fact extraction and {VER}ification}.
\newblock In \emph{Proceedings of the 2018 Conference of the North {A}merican Chapter of the Association for Computational Linguistics: Human Language Technologies, Volume 1 (Long Papers)}, pages 809--819, New Orleans, Louisiana. Association for Computational Linguistics.

\bibitem[{Tian et~al.(2024)Tian, Mitchell, Yao, Manning, and Finn}]{tianfine}
Katherine Tian, Eric Mitchell, Huaxiu Yao, Christopher~D Manning, and Chelsea Finn. 2024.
\newblock Fine-tuning language models for factuality.
\newblock In \emph{The Twelfth International Conference on Learning Representations}.

\bibitem[{Touvron et~al.(2023)Touvron, Lavril, Izacard, Martinet, Lachaux, Lacroix, Rozi{\`e}re, Goyal, Hambro, Azhar et~al.}]{touvron2023llama}
Hugo Touvron, Thibaut Lavril, Gautier Izacard, Xavier Martinet, Marie-Anne Lachaux, Timoth{\'e}e Lacroix, Baptiste Rozi{\`e}re, Naman Goyal, Eric Hambro, Faisal Azhar, et~al. 2023.
\newblock Llama: Open and efficient foundation language models.
\newblock \emph{arXiv preprint arXiv:2302.13971}.

\bibitem[{Wright et~al.(2022)Wright, Wadden, Lo, Kuehl, Cohan, Augenstein, and Wang}]{wright-etal-2022-generating}
Dustin Wright, David Wadden, Kyle Lo, Bailey Kuehl, Arman Cohan, Isabelle Augenstein, and Lucy~Lu Wang. 2022.
\newblock \href {https://doi.org/10.18653/v1/2022.acl-long.175} {Generating scientific claims for zero-shot scientific fact checking}.
\newblock In \emph{Proceedings of the 60th Annual Meeting of the Association for Computational Linguistics (Volume 1: Long Papers)}, pages 2448--2460, Dublin, Ireland. Association for Computational Linguistics.

\bibitem[{Xu et~al.(2024{\natexlab{a}})Xu, Sun, Zheng, Geng, Zhao, Feng, Tao, Lin, and Jiang}]{xu2024wizardlm}
Can Xu, Qingfeng Sun, Kai Zheng, Xiubo Geng, Pu~Zhao, Jiazhan Feng, Chongyang Tao, Qingwei Lin, and Daxin Jiang. 2024{\natexlab{a}}.
\newblock \href {https://openreview.net/forum?id=CfXh93NDgH} {Wizard{LM}: Empowering large pre-trained language models to follow complex instructions}.
\newblock In \emph{The Twelfth International Conference on Learning Representations}.

\bibitem[{Xu et~al.(2024{\natexlab{b}})Xu, Jiang, Niu, Deng, Poovendran, Choi, and Lin}]{xu2024magpie}
Zhangchen Xu, Fengqing Jiang, Luyao Niu, Yuntian Deng, Radha Poovendran, Yejin Choi, and Bill~Yuchen Lin. 2024{\natexlab{b}}.
\newblock \href {https://api.semanticscholar.org/CorpusID:270391432} {Magpie: Alignment data synthesis from scratch by prompting aligned llms with nothing}.
\newblock \emph{ArXiv}, abs/2406.08464.

\bibitem[{Yu et~al.(2024)Yu, Jiang, Shi, YU, Liu, Zhang, Kwok, Li, Weller, and Liu}]{yu2024metamath}
Longhui Yu, Weisen Jiang, Han Shi, Jincheng YU, Zhengying Liu, Yu~Zhang, James Kwok, Zhenguo Li, Adrian Weller, and Weiyang Liu. 2024.
\newblock \href {https://openreview.net/forum?id=N8N0hgNDRt} {Metamath: Bootstrap your own mathematical questions for large language models}.
\newblock In \emph{The Twelfth International Conference on Learning Representations}.

\bibitem[{Zha et~al.(2023)Zha, Yang, Li, and Hu}]{zha-etal-2023-alignscore}
Yuheng Zha, Yichi Yang, Ruichen Li, and Zhiting Hu. 2023.
\newblock \href {https://aclanthology.org/2023.acl-long.634} {{A}lign{S}core: Evaluating factual consistency with a unified alignment function}.
\newblock In \emph{Proceedings of the 61st Annual Meeting of the Association for Computational Linguistics (Volume 1: Long Papers)}, pages 11328--11348, Toronto, Canada. Association for Computational Linguistics.

\end{thebibliography}

\clearpage

\appendix

\section{Appendix}

\subsection{Generation Examples}\label{app:generation_examples}

\begin{table}[ht!]
\small
\begin{tabular}{p{0.95\linewidth}}
\toprule
\textbf{Source:} El nevado, que es el punto central del Parque Nacional Natural Nevado del Huila, está dividido territorialmente entre los departamentos de Huila, Tolima y Cauca, siendo su cima el punto más elevado de los tres. \\
\vspace{1mm}
\textbf{EN:} \textit{The snow-capped mountain, which is the central point of the Nevado del Huila National Natural Park, is territorially divided between the departments of Huila, Tolima and Cauca, its summit being the highest point of the three.} \\
\midrule
\textbf{Supports:} El Nevado del Huila es territorialmente dividido entre los departamentos de Huila, Tolima y Cauca.  \\
\vspace{1mm}
\textbf{EN:} \textit{The Nevado del Huila is territorially divided between the departments of Huila, Tolima and Cauca.}\\
\midrule
\textbf{Refutes}: La cima del Nevado del Huila es la puntalina más baja de los tres departamentos en que se encuentra. \\
\vspace{1mm}
\textbf{EN:} \textit{The summit of Nevado del Huila is the lowest point of the three departments in which it is located.}\\
\midrule
\textbf{Not-info:} El Nevado del Huila es el punto más alto entre los puntos elevados de la Cordillera Central en Colombia. \\
\vspace{1mm}
\textbf{EN:} \textit{The Nevado del Huila is the highest point among the high points of the Cordillera Central in Colombia.}\\
\midrule
\textbf{Topic:} Nevado del Huila\\
\textbf{EN:} \textit{Nevado del Huila}\\
\bottomrule
\end{tabular}
\caption{Generation example of Spanish claims for each class.}
  \label{table:example_claims_es}
\end{table}

\begin{table}[ht!]
\small
\begin{tabular}{p{0.95\linewidth}}
\toprule
\textbf{Source:} Durch den Britisch-Niederländischen Vertrag von 1814 fiel Berbice an Großbritannien. \\
\vspace{1mm}
\textbf{EN:} \textit{Berbice fell to Great Britain as a result of the British-Dutch Treaty of 1814.} \\
\midrule
\textbf{Supports:} Berbice ist nach dem Vertrag von 1814 zu GroÃŸbritannien gefallen.  \\
\vspace{1mm}
\textbf{EN:} \textit{Berbice fell to Great Britain after the Treaty of 1814.}\\
\midrule
\textbf{Refutes}: Berbice wurde 1814 durch den Britisch-Niederländischen Vertrag an die Niederlande zurückgegeben. \\
\vspace{1mm}
\textbf{EN:} \textit{Berbice was returned to the Netherlands in 1814 by the British-Dutch Treaty.}\\
\midrule
\textbf{Not-info:} Die Menge von Niederländisch-Berbice-Siedlungen war größer als die der britischen Siedlungen, bevor Berbice an Großbritannien fiel. \\
\vspace{1mm}
\textbf{EN:} \textit{The number of Dutch Berbice settlements was greater than that of the British settlements before Berbice fell to Great Britain.}\\
\midrule
\textbf{Topic:} Berbice-Niederländisch\\
\textbf{EN:} \textit{Berbice-Nederlands}\\
\bottomrule
\end{tabular}
\caption{Generation example of German claims for each class.}
  \label{table:example_claims_de}
\end{table}

\subsection{Generation Prompts}\label{app:generation_prompts}
We present the prompts for generating \textit{refuting} and \textit{not-info} claims in Tables \ref{table:negated_prompt} and \ref{table:notinfo_prompt}.

\begin{table*}[t!]
\centering
\begin{tabularx}{\textwidth}{|X|}
\hline
Act as an expert in generating claims in <language>. I will give you one sentence in {language}, about the topic: "<topic>". The sentence is: "<sources>". Generate a single short and falsified claim in {language} based on the information about the sentence I provide. Try to use comparative and superlative adjectives between objects (e.g., larger, smaller, more, faster, higher), while ensuring that the sentence does not support the claim. Do not improvise. Use only the information I provide. Do not add any explanation, refrain from adding extra information nor your opinion on whether the claim is true or not, as long as it is not supported by the sentence. The claim should have less than 30 words. Do not make any reference to the sentence in your answer, make it self-contained. Write your answer in <language> only. Evaluate at the end how good the generated claim is. Your response must be in the same format as the JSON in the examples below.  \\

Your response must be in the same format as the JSON in the examples below.   \\
\{\{      \\
\hspace{10mm} "CLAIM": "Write a single short and falsified claim in <language>, based on the sentence",   \\
\hspace{10mm} "SELF-CONTAINED": "Assess how self-contained the generated claim is on a scale of 1 to 5.",   \\
\hspace{10mm} "CATEGORY": "Categorise whether the given sentence not supported (C0), supported (C1) the generated claim (independently of actual veracity of the claim), or not verifiable (C2)",   \\
\hspace{10mm} "SUPPORTED BY ORIGINAL SENTENCE": "Assess how supported the claim is by the original sentence on a scale of 1 to 5.",   \\
\hspace{10mm} "FACTUAL": "Assess how factual the claim is based on the original sentence [real/non-fiction/non-fantastic]",   \\
\hspace{10mm} "OBJECTIVE": "Assess how objective the claim is on a scale of 1 to 5",   \\
\hspace{10mm} "OVERALL QUALITY": "Assess the overall quality of the claim on a scale of 1 to 5"  \\
\}\}. \\
\hline
\end{tabularx}
\caption{The prompt for generating \textit{refuting} claims.}
  \label{table:negated_prompt}
\end{table*}

\begin{table*}[th!]
\centering
\begin{tabularx}{\textwidth}{|X|}
\hline
Act as an expert in generating claims in <language>. I will give you one evidence in <language>, about the topic: "<topic>". Generate a single short and specific claim in <language>, relevant to the information about the evidence I will provide. The claim should be not verifiable based on the evidence provided. Also, the claim should use comparative form (e.g., larger, smaller, more, faster, higher) to make comparisons between objects, people, ideas, dates or numbers in the evidence. Do not add any explanation, refrain from adding extra information nor your opinion on whether the claim is true or not. The claim should have less than 30 words. The evidence is: "<sources>". Do not make any reference to the sentence in your answer, make it self-contained. Write your answer in <language> only. Evaluate at the end how good the generated claim is. Your response must be in the same format as the JSON in the examples below.  \\

Your response must be in the same format as the JSON in the examples below.   \\
\{\{      \\
\hspace{10mm} "CLAIM": "Write a single short and falsified claim in <language>, based on the sentence",   \\
\hspace{10mm} "SELF-CONTAINED": "Assess how self-contained the generated claim is on a scale of 1 to 5.",   \\
\hspace{10mm} "CATEGORY": "Categorise whether the given sentence not supported (C0), supported (C1) the generated claim (independently of actual veracity of the claim), or not verifiable (C2)",   \\
\hspace{10mm} "SUPPORTED BY ORIGINAL SENTENCE": "Assess how supported the claim is by the original sentence on a scale of 1 to 5.",   \\
\hspace{10mm} "FACTUAL": "Assess how factual the claim is based on the original sentence [real/non-fiction/non-fantastic]",   \\
\hspace{10mm} "OBJECTIVE": "Assess how objective the claim is on a scale of 1 to 5",   \\
\hspace{10mm} "OVERALL QUALITY": "Assess the overall quality of the claim on a scale of 1 to 5"  \\
\}\}. \\
\hline
\end{tabularx}
\caption{The prompt for generating \textit{not-info} claims.}
  \label{table:notinfo_prompt}
\end{table*}

\subsection{Dataset Statistics}\label{app:data_statistics}
Table \ref{table:statistics_results1} presents the statistics of our datasets without MNLI filtering using automatic metrics.

\begin{table*}[!ht]
    \small
    \centering
    \resizebox{\textwidth}{!}{
    \begin{tabular}{lrrrrrrrrr}
    \toprule
                         & \multicolumn{2}{c}{Statistics}  & \multicolumn{4}{c}{LLM evaluation} & \multicolumn{3}{c}{Lexical similarity}\\
                           \cmidrule(lr){2-3}                \cmidrule(lr){4-7}      \cmidrule(lr){8-10}
          ~              &  words$_\mu$   & words$_{sd}$& self-contained &  support & objective &  quality &  BLEU-4 & ROUGEL & METEOR\\              
    \midrule
    EN                   &                &             &                &         &           &        &         &        &     \\
    \hspace{3mm} supports &     16         &     6       &     5          &   5     &    5      &   4    & 0.21  & 0.47  & 0.47 \\
    \hspace{3mm} refutes &    13         &  4          &  5             &   1     &   4       &  4     & 0.10  & 0.34 & 0.31 \\
    \hspace{3mm} not-info &    15         &   4         &   5            &   2     &   4       &  4     & 0.06  & 0.24 & 0.24 \\
    \midrule
    ES                   &                  &              &                &         &           &         &         &        &    \\
    \hspace{3mm} supports &     16        &     6     &    5        &  5    &  5     &  4    & 0.19   & 0.47 & 0.43\\
    \hspace{3mm} refutes &    13          &   4         &       5      &    1   &    4     &  4    &  0.08  & 0.31   & 0.27 \\
    \hspace{3mm} not-info &    14          &   4         &   5           &   1    &   4      & 4      & 0.05  & 0.24 & 0.20 \\
    \midrule
    DE                   &                &             &               &        &          &        &     &    &  \\
    \hspace{3mm} supports &    13         &     5       &     5        &  5     &    5     &   4    &  0.20  & 0.46  & 0.43 \\
    \hspace{3mm} refutes &   10           &   4         &   5          &  1     &   4      &  4     & 0.10  & 0.31  & 0.28 \\
    \hspace{3mm} not-info &   11           &   4         &   5           &  1     &  4       & 4      & 0.05  & 0.21 & 0.18 \\
    \bottomrule
    \end{tabular}}
    \caption{Statistics and evaluation of our dataset without MNLI filtering.  Columns $words_\mu$ and $words_\text{sd}$ show the average and standard deviation of the number of words per claim. The following columns show average values for the respective metric denoted as the column names. Lexical similarity metrics are computed between sources and generated claims.}
    \label{table:statistics_results1}
\end{table*}

\subsection{Test Datasets Pre-Processing}\label{app:datasets_pre}
The processing of VitaminC (extracted from HuggingFace\footnote{https://huggingface.co/datasets/tals/vitaminc}) is direct. We take the `evidence' field as the source and maintain the classes `SUPPORTS', `REFUTES' and `NOT ENOUGH INFORMATION' as in our pipeline.

X-Fact dataset is also obtained from HuggingFace\footnote{https://huggingface.co/datasets/utahnlp/x-fact}. We reorganise its original classes in three regarding our setup: `true', `partly true/misleading' and `mostly true' are relabeled as `supports'; `false' is redirected to the `refutes' class; and the rest is considered `not-info'. Regarding the source, we concatenate all evidence fields (usually from 1 to 5) found on each claim except those containing the text `<DUMMY\_EVIDENCE>', as a unique input.

NLI-Fever, also found in HugginFace\footnote{https://huggingface.co/datasets/pietrolesci/nli\_fever}, has a different distribution of classes regarding the NLI style. We frame `entailment' as `supports', `contradiction' as `refutes' and `neutral' as `not-info'. The premise is selected as the claim and the hypothesis is considered the evidence source.

\subsection{Model Training}\label{app:training_details}
We adopt a learning rate of 1e-6, a warm-up proportion of 0.06, Adam epsilon of 1e-6 and an accumulated gradient batch size of 2. To prevent overfitting, we included early stopping with a patience of 2 epochs. Regarding time executions, the longest training finishes with 9 epochs and takes 25 hours.

\subsection{Generation Prompt for Cross-model and cross-language evaluation}\label{app:Cross-model-eval}

We include in Table~\ref{table:fact_checking_prompt} the prompt used to evaluate fact-checking using LLMs shown in Tables~\ref{tab:cross_model} and~\ref{tab:x_fact_languages}.

\begin{table*}[th!]
\centering
\begin{tabularx}{\textwidth}{|X|}
\hline
\vspace{0.1cm}
 [\emph{role}: system] \\ You are a helpful assistant and an expert fact checker. You will receive a claim and some source text. You have to determine what information the source contains with respect to the claim, independently of common knowledge. If it is not relevant to the truth value of the claim, label it as not-info or irrelevant (I). If it is relevant to the truth value of the claim, label it as true or supporting the claim (S), or false or contradicting the claim (C). Write first the label as 'Label: C' for example. Don't give any other explanation, only the label.\\ \\
 
 [\emph{role}: user]\\ Claim: <claim>\\
 Evidence: <evidence> \\ \\
\hline
\end{tabularx}
\caption{A prompt example for fact-checking using an LLM.}
  \label{table:fact_checking_prompt}
\end{table*}

\subsection{Transformer Comparison.}\label{app:transformer_comparison}

Table~\ref{tab:models} presents results for fine-tuning three transformer-based models on the synthetic dataset: \textbf{RoBERTa-large}, \textbf{DeBERTa-large} and \textbf{mDeBERTa-base}. While \textit{DeBERTa-large} achieves the highest F1 scores, we opted for \textit{mDeBERT-base} due to its multilingual capabilities and smaller size, which enables faster experimentation. We hypothesize that an ideal model for multilingual fact-checking would be a DeBERTa-large-sized model pre-trained on a diverse multilingual corpus.

\begin{table*}[h]
\small
\centering
\begin{tabular}{l
>{\columncolor[HTML]{DAE9F8}}l 
>{\columncolor[HTML]{F2CEEF}}l 
>{\columncolor[HTML]{FBE2D5}}l }
\toprule
\multicolumn{1}{c}{\cellcolor[HTML]{D0D0D0}\textbf{Language}} & \cellcolor[HTML]{D0D0D0}\textbf{mdeberta-base} & \cellcolor[HTML]{D0D0D0}\textbf{deberta-large} & \cellcolor[HTML]{D0D0D0}\textbf{roberta-large} \\
\midrule
\textbf{en}                                                            & 0.946                                          & \textbf{0.963}                                 & 0.931                                          \\
\textbf{es}                                                            & 0.932                                          & \textbf{0.946}                                 & 0.906                                          \\
\textbf{de}                                                            & 0.931                                          & \textbf{0.943}                                 & 0.904                                          \\
\textbf{MEAN} (STD)                                                          & 0.937 (0.008)                                         & \textbf{0.951} (0.011)                                & 0.914 (0.015)     \\              \bottomrule                   
\end{tabular}
\caption{Macro F1 scores when training and testing in our synthetic dataset for each language and different base models. Last row shows the mean and the standard deviation, in parenthesis, across all languages.}
\label{tab:models}
\end{table*}

\subsection{Ablation studies}
\label{app:ablation}

To understand the impact of applying MNLI filtering in data generation, we conduct ablation studies training the \textit{mDeBERTa-v3-base} model on VitaminC datasets, incorporating our synthetic data without MNLI filtering (no-MNLI). 
As shown in Table \ref{tab:ablation}, the models trained with MNLI filtering surpass the ones trained without MNLI filtering for all scenarios except for English. This suggests employing MNLI filtering is beneficial for curating better-quality multilingual fact-checking datasets.

\begin{table}[h!]
\small
\centering
\begin{tabular}{cl
>{\columncolor[HTML]{F2CEEF}}l 
>{\columncolor[HTML]{DAF2D0}}l }
\toprule
\cellcolor[HTML]{D0D0D0}\textbf{Data} & \multicolumn{1}{c}{\cellcolor[HTML]{D0D0D0}\textbf{lang}} & \cellcolor[HTML]{D0D0D0}\textbf{MNLI} & \cellcolor[HTML]{D0D0D0}\textbf{no-MNLI} \\
\midrule
                                      & en                                                            & 0.846                  & \textbf{0.847}                              \\
                                      & es                                                            & \textbf{0.825}                  & 0.818                              \\
\multirow{-3}{*}{VitaminC}            & de                                                            & \textbf{0.819}                           & 0.815                    \\
              
\bottomrule
\end{tabular}
\caption{Ablation studies showing the impact of MNLI filtering component. Results show macro F1 scores in VitaminC dataset across different languages.}\label{tab:ablation}
\end{table}

\end{document}